\documentclass{article}

\usepackage{times}
\usepackage{graphicx} % more modern
\usepackage{subfigure}
\usepackage{caption}
\usepackage{natbib}

\usepackage{algorithm}
\usepackage{algorithmic}

\usepackage[pdftex, colorlinks]{hyperref}
\usepackage[numbered]{bookmark}
\usepackage{tabularx}
\usepackage{amsmath,amssymb} % define this before the line numbering.
\usepackage{multirow}
\usepackage{varwidth}
\usepackage{color}
\usepackage{listings}
\usepackage{enumitem}

\usepackage[accepted]{icml2016}

\icmltitlerunning{Multi-Bias Non-linear Activation in Deep Neural Networks}

\begin{document}

\twocolumn[
\icmltitle{Multi-Bias Non-linear Activation in Deep Neural Networks}
% on the importance of bias-guided non-linearity learning

\icmlauthor{Hongyang Li}{yangli@ee.cuhk.edu.hk}
\icmlauthor{Wanli Ouyang}{wlouyang@ee.cuhk.edu.hk}
\icmlauthor{Xiaogang Wang}{xgwang@ee.cuhk.edu.hk}
\icmladdress{Department of Electronic Engineering, The Chinese University of Hong Kong.}

%\icmladdress{Their Fantastic Institute, 27182 Exp St., Toronto, ON M6H 2T1 CANADA}

\icmlkeywords{boring formatting information, machine learning, ICML}

\vskip 0.3in
]

\begin{abstract}
As a widely used non-linear activation, Rectified Linear Unit (ReLU)  separates noise and signal in a feature map by learning a threshold or bias. However, we argue that the classification of noise and signal not only depends on the magnitude of responses, but also the context of how the feature responses would be used to detect more abstract patterns in higher layers. In order to output multiple response maps  with magnitude in different ranges for a particular visual pattern, existing networks employing ReLU and its variants have to learn a large number of redundant filters.
		%ReLU and its variants only output one feature map by taking a feature map from a convolution layer as input.
		%In order to output multiple response maps of visual pattern with responses in different ranges,
		
In this paper, we propose a multi-bias non-linear activation (MBA) layer to explore the information hidden in the magnitudes of responses. It is placed after the convolution layer to decouple the responses to a convolution kernel into multiple maps by multi-thresholding magnitudes, thus generating more patterns in the feature space at a low computational cost.
%in both the number of parameters and computation.
It provides great flexibility of selecting responses to different visual patterns in different magnitude ranges to form rich representations in higher layers.
		%
    %Therefore, the filter learned not only contains
    %the distinct magintude range to the same signal within each channel by adding more non-linearity, but also encodes a large variety of pattern representations by considering the magnitudes across channels jointly.
    %
Such a simple and yet effective scheme achieves the state-of-the-art performance %in image classification 
%on CIFAR-10 (5.38\%) and CIFAR-100 (24.1\%).
on several benchmarks.
\end{abstract}

\section{Introduction}\label{intro}

Deep neural networks \cite{alexnet,hinton_nature,hinton_science} has made great progress on different domains and applications in recent years.
The community has witnessed the machine trained with deep networks and massive data being the first computer program defeating a European professional in the game of Go \cite{play_go_nature};
the convolutional neural network surpassing human-level performance in image classification \cite{prelu};
%the end-to-end deep learning system with HPC techniques recognizing language in English and Mandarin \cite{deep_speech},
the deep neural network framework to build an acoustic model in speech recognition \cite{deepSpeechReviewSPM2012}.
%etc.

\begin{figure}
	\centering
	\includegraphics[width=\linewidth]{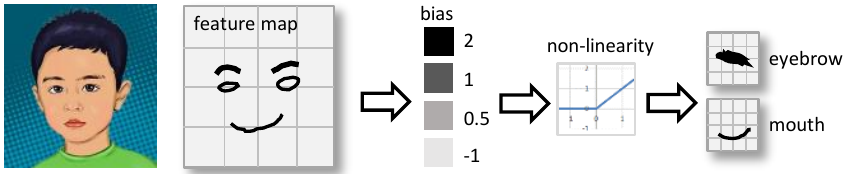}\\
	\vspace{-.1cm}
	\caption{Example illustrating how biases can select different patterns. The eyes and mouth have different magnitudes in their responses to the curved edges. Whether responses of various strength should be considered as informative signal or noise depends on more abstract patterns to be detected in higher layers. By adding different biases to the map with MBA, they are strengthened, preserved, or eliminated according to each bias.}\label{example}
\end{figure}

The importance of activation function has been recognized in the design of deep models. It not only fits complex data distributions but also achieves important invariance to various noise, data corruption and transforms affecting recognition \cite{Bruna}. It is placed after every convolution layer with sigmoid or hyperbolic tangent as non-linear activation. If the input responses are too large positively or negatively, they are compressed to a saturation value through the nonlinear mapping and thus invariance is achieved. The rectified linear unit (ReLU) %, \emph{i.e.}, $y = max(0,x)$, 
is found to be particularly effective \cite{relu_icml10, alexnet} in deep neural networks and widely used. It separates noisy signals and informative signals in a feature map by learning a threshold (bias). Certain amount of information is discarded after the non-linear activation. However, ReLU also has limitation because of the observations.
%from us.

Given the same convolution kernel in the convolution layer, we observe that different magnitudes of responses may indicate different patterns.  
An illustrative example is shown in Figure \ref{example}. The eyes should have higher magnitudes in the responses to curved edges compared with those on the mouth because edge contrast on eyes is generally higher than that on mouths. Therefore, it is desirable to separate the responses according to its magnitude.

More importantly, the separation between informative signal and noise not only depends on the magnitudes of responses, but also the context of how the feature responses would be used to detect more abstract patterns in higher layers. In the hierarchical representations of deep neural networks, a pattern detected at the current layer serves as a sub-pattern to be combined into a more abstract pattern in its subsequent layers. For example, curved edges in Figure \ref{example} are detected in the current layer and one of filters in its subsequent layer detects eyes. It requires high response of curved edges and treat the response with moderate magnitude as noise. However, for another filter in its subsequent layer to detect mouth, moderate responses are enough. %Therefore, separation of responses according to its magnitude helps the subsequent layer to choose the required patterns accordingly.

Unfortunately, if feature responses with ReLU module are removed by thresholding, they cannot be recovered. A single thresholding cannot serve for multiple purposes. In order to output multiple response maps with magnitudes in different ranges for a particular visual pattern, networks employing ReLU have to learn multiple redundant filters. A set of redundant filters learn similar convolution kernels but distinct bias terms. It unnecessarily increases the computation cost and model complexity, and is easier to overfit training data.

Many delicate activation functions have been proposed to increase the flexibility of nonlinear function.
Parametric ReLU \cite{prelu} generalizes Leaky ReLU \cite{leaky_relu} by learning the slope of the negative input, which yields an impressive learning behavior on large-scale image classification benchmark.
 Other variants in the ReLU family include Randomized Leaky ReLU \cite{leacky_relu} where the slope of the negative input is randomly sampled, and Exponential Linear Unit \cite{iclr16_elu} which has an exponential shape in the negative part and ensures a noise-robust deactivation state.
%
%In these approaches, the output value of the activation function only depends on the magnitude of the response obtained by convolution. The non-linear activation function is a key in discarding noisy signals while keeping informative signals.  However, these designs are not able to decouple the information hidden in the magnitude of response in a efficient way.
Although these variants can reweight feature responses whose magnitudes are in different ranges, they cannot separate them into different feature maps. 

\begin{figure}[t]	
	\includegraphics[width=\linewidth]{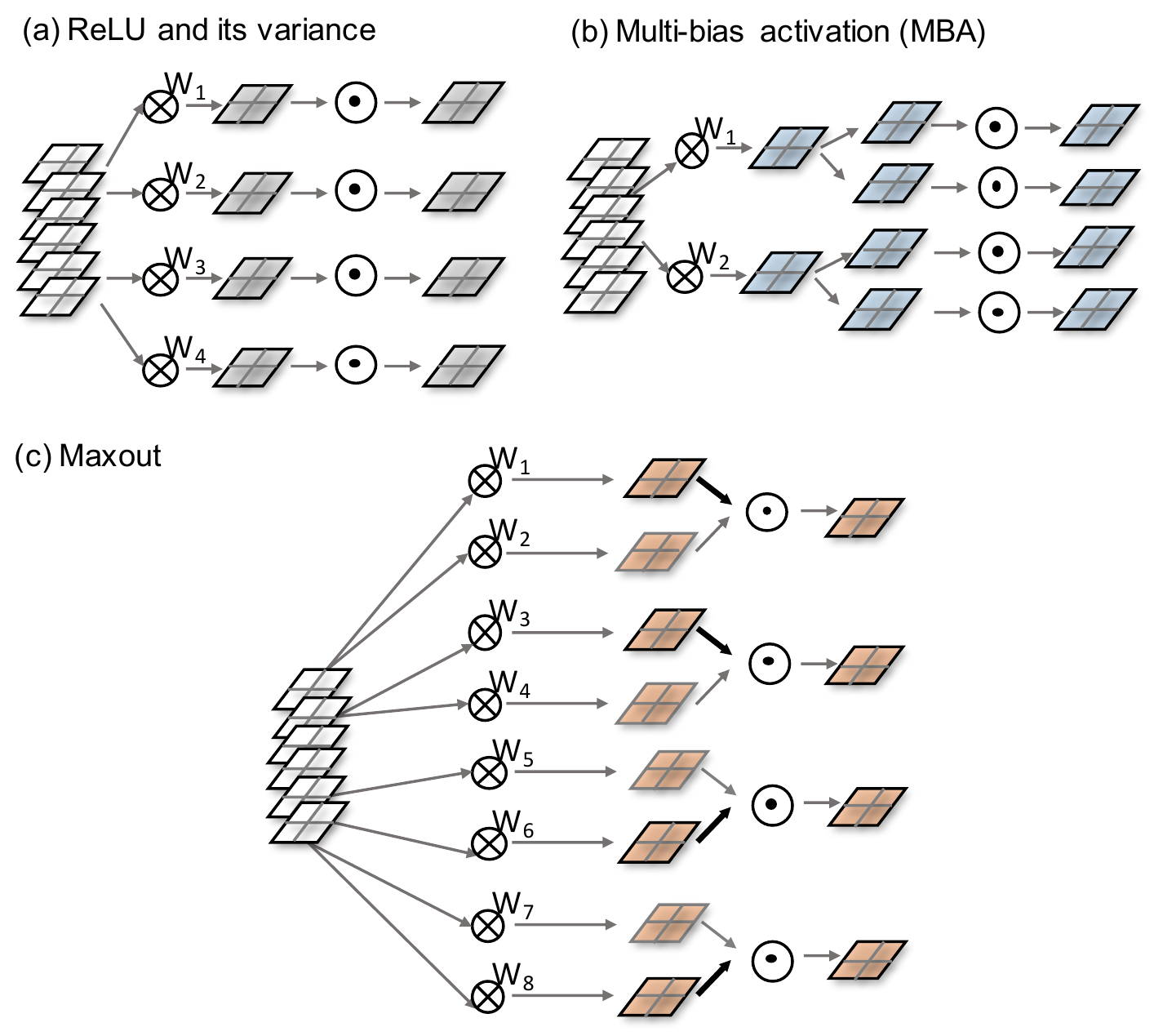}
	\caption{A comparison of MBA with ReLU and its variants, and maxout. (a) Given a feature map as input, ReLU and its variants output a single feature map. In order to decouple the responses to a visual pattern to multiple magnitude ranges,  the network has to learn multiple filters which have similar convolution kernels $\{\textbf{W}_k\}$ and distinct biases. (b) MBA takes one feature map as input and output multiple band maps by introducing multiple biases. It does not need to learn redundant convolution kernels. The decoupling of signal strength increase the expressive power of the net. (c) Maxout combines multiple input feature maps to a single output map. $\otimes$ denotes convolution operation and $\odot$ denotes non-linear activation. }\label{fig:overall}
	\vspace{-.2cm}
\end{figure}

As summarized in Figure \ref{fig:overall}, given a feature map as input, non-linear activation ReLU and its variants only output a single feature map.
However, the MBA module outputs multiple feature maps without having to learn as many kernels as does ReLU.
In some sense, our idea is opposite to the maxout network \cite{maxout}. Maxout is also a non-linear activation. However, given $K$ feature maps generated by $K$ convolution kernels, it combines them to a single feature map:
\begin{equation*}\label{}
h(\mathbf{x}) = \arg \max_{i\in\{1,\cdots,K  \}} \mathbf{w}_i \mathbf{x} + \mathbf{b}_i,
\end{equation*}
where $\mathbf{x}\in\mathbb{R}^d$ is an image, $\mathbf{w}_i\in \mathbb{R}^{d\times p \times K}$ is a convolution kernel, and $b_i\in \mathbb{R}^{p \times K}$ is a bias.
The motivations of MBA and maxout are different and can be jointly used in the network design to balance the number of feature maps.
%By incorporating multiple linear functions, the network has more expressive power to approximate any continuous function. Our work decouples single channel while the maxout network combines multiple channels. Therefore, our work is complementary to the maxout network.

To this end, we propose a multi-bias non-linear activation (MBA) layer for deep neural networks. 
It decouples a feature map obtained from a convolution kernel to multiple maps, called \textit{band maps},
according to the magnitudes of responses. %without increasing the model complexity significantly. 
This is implemented by introducing different biases, which share the \textit{same} convolution kernel, 
imposed on the feature maps and then followed by the standard ReLU. 
Each decoupled band map corresponds to a range in the response magnitudes to a convolution kernel, and the range is learned. 
The responses in different magnitude ranges in the current layer are selected and combined in a flexible way by each filter in the subsequent layer.
%Specifically, we add \emph{bias} layer after each feature map of the current convolutional layer and the number of feature maps could be increased by enforcing different bias thresholds on them. After passing through the non-linearity unit, we directly forward these across-channel maps into the next layer.
We provide analysis on the effect of the MBA module when taking its subsequent layer into account. 
Moreover,
it is shown that the piece-wise linear activation function \cite{piecewise_linear} is a special case of MBA, where MBA provides more flexibility in decoupling the magnitudes and also in combining band maps from different convolution kernels. 
Finally,
The experimental results on the CIFAR and SVHN datasets show that such a simple and yet effective algorithm can achieve state-of-the-art performance. 
%Codes and results will be available upon acceptance.

%In summary, in this work, we raise an important issue regarding to ReLU and its variants on separating noise and informative signal. A new multi-bias non-linear activation (MBA) is proposed.
%%\footnote{We use the term MBA module or bias model alternatively to refer to our algorithm in the following context.}.
%In order to separate detection signals of a visual pattern at different strength levels, existing deep networks have to employ a large number of filters with redundancy on convolution kernels. MBA can effectively reduce such redundancy while increasing the expressive power of the network. It proves that the piece-wise linear activation function is a special case of multi-bias activation, but with much less expressive power.

%%%%%%%%%%%%%%%%%%%
%The rest of the paper is arranged as follows.
%Sec.\ref{alg} describes the details on the multi-bias non-linear activation function.
%%
%In particular, we point out the significant difference between  our algorithm and the piecewise linear unit in Sec.\ref{comparison}; thorough experiments on  component analysis and comparison with the state-of-the-arts are provided in Sec.\ref{exp}; finally, the concluding remarks are covered in Sec.\ref{conclusion}.
%%%%%%%%%%%%%%%%%%%%%%%%%%%%%

\begin{figure*}
	\centering
	\includegraphics[width=\linewidth]{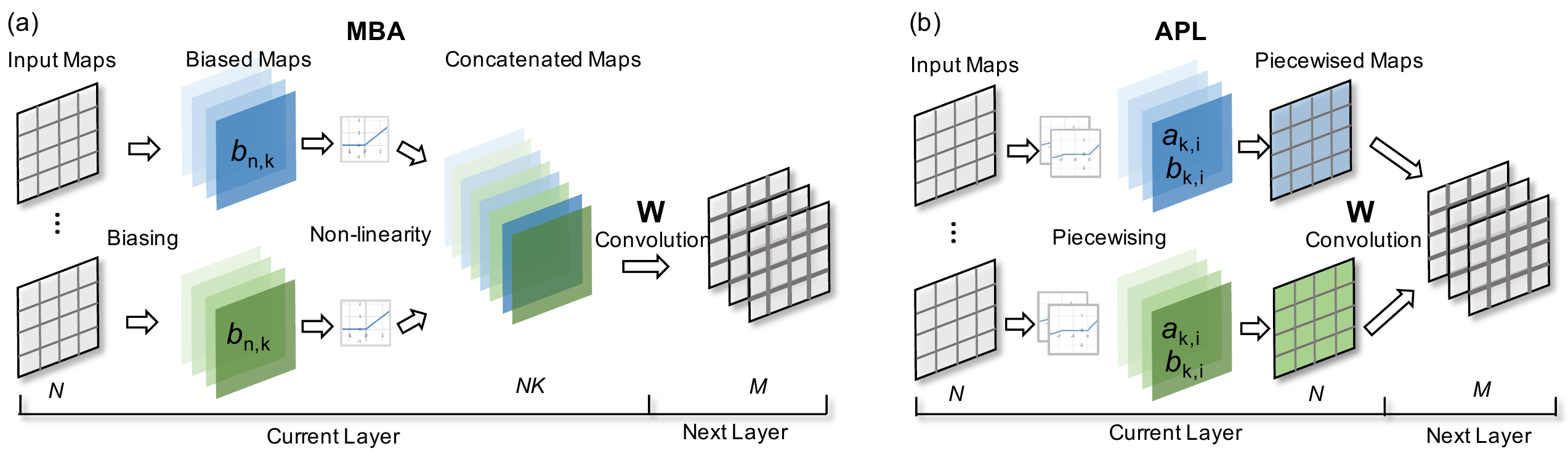}\\
	\vspace{-.15cm}
	\caption{(a) The proposed multi-bias activation (MBA) model. Given the input feature maps, the MBA module adds biases $b_{n,k}$ on these maps to generate $NK$ band maps, then these `biasing' maps are fed into the subsequent convolutional layer in a flexible way.
		(b) The piecewise linear function (APL) module where a set of $K$ learnable parameters $\{a_k, b_k\}$
		sums up the maps within each channel before feeding $N$ output maps into the next convolution layer and thus providing no cross-channel information.
		The additional parameters brought by these two modules are $NK$ and $2KWH$, respectively.		
	}\label{fig:bias}
\end{figure*}

\section{Multi-bias Non-linear Activation }\label{alg}
%%% my revised version
The goal of this work is to decouple a feature map into multiple band maps by introducing a MBA layer, thus enforcing different thresholds on the same signal
where in some cases responses in a certain range carry useful patterns and in another case they are merely noise. After passing through ReLU, these band maps are selected
and combined by the filters in the subsequent convolution to represent more abstract visual patterns with large diversity.

\subsection{Model formulation}\label{sec:model_formulation}
Fig.\ref{fig:bias} (a) depicts the pipeline of the MBA module.
After convolution in CNN, we obtain a set of feature maps.
We represent the $n$-th feature map by vector $\mathbf{x}_{n} \in \mathbb{R}^{W H}$, where $n= 1\ldots N$, $W$ and $H$ denote the spatial width and height of the feature map, respectively. If the input of the MBA layer is the response obtained by a fully-connected layer, we can simply treat $W=H=1$.
%\emph{e.g.}, the output feature maps of the \emph{current} convolutional layer,
The MBA layer separates $\mathbf{x}_{n}$ into  $K$ feature maps $\hat{\mathbf{x}}_{n,k}$ as follows:
\begin{align}\label{eqn:bias}
\begin{split}
 \hat{\mathbf{x}}_{n,k} =  \sigma(\mathbf{x}_n + b_{n,k}) \textrm{ for } k=1, \ldots, K, \\
  \end{split}
\end{align}
where $\sigma(\cdot)$ is the element-wise nonlinear function. 
Note that the only parameter introduced by the MBA module is a scalar $b_{n,k}$. %of dimension $WH$.
Denote $x_{n,i}$ and $\hat{x}_{n,k,i}$ as the $i$-th element in the map $\mathbf{x}_n$ and  $\hat{\mathbf{x}}_{n,k}$ respectively, 
where $i=1, \dots, WH$, 
we have the element-wise output form of the MBA module defined as:
\begin{align}\label{eqn:bias_ele2}
\begin{split}
\hat{x}_{n,i}=\sigma( x_{n,i} +b_{n,k}).
  \end{split}
\end{align}
In this paper, we mainly consider using ReLU
%$max\{0, \cdot\}$
as the nonlinear function %$\sigma$ 
because it is found to be successful in many applications.  In ReLU, the response  $x_{n, i}$  is thresholded by the bias $b_{n,k}$ as follows:
\begin{align}\label{eqn:bias_ele}
\begin{split}
& \textrm{if } x_{n,i} \leq -b_{n,k}, \hspace{0.1in} \textrm{then }\hat{x}_{n,i}=0, \\
&\textrm{if } x_{n,i} > -b_{n,k}, \hspace{0.1in} \textrm{then }\hat{x}_{n,i}=x_{n,i} +b_{n,k}.
  \end{split}
\end{align}

\begin{figure*}[t]
	\footnotesize
	\begin{minipage}[b]{0.24\linewidth}
		\centering
		\includegraphics[width=1\linewidth]{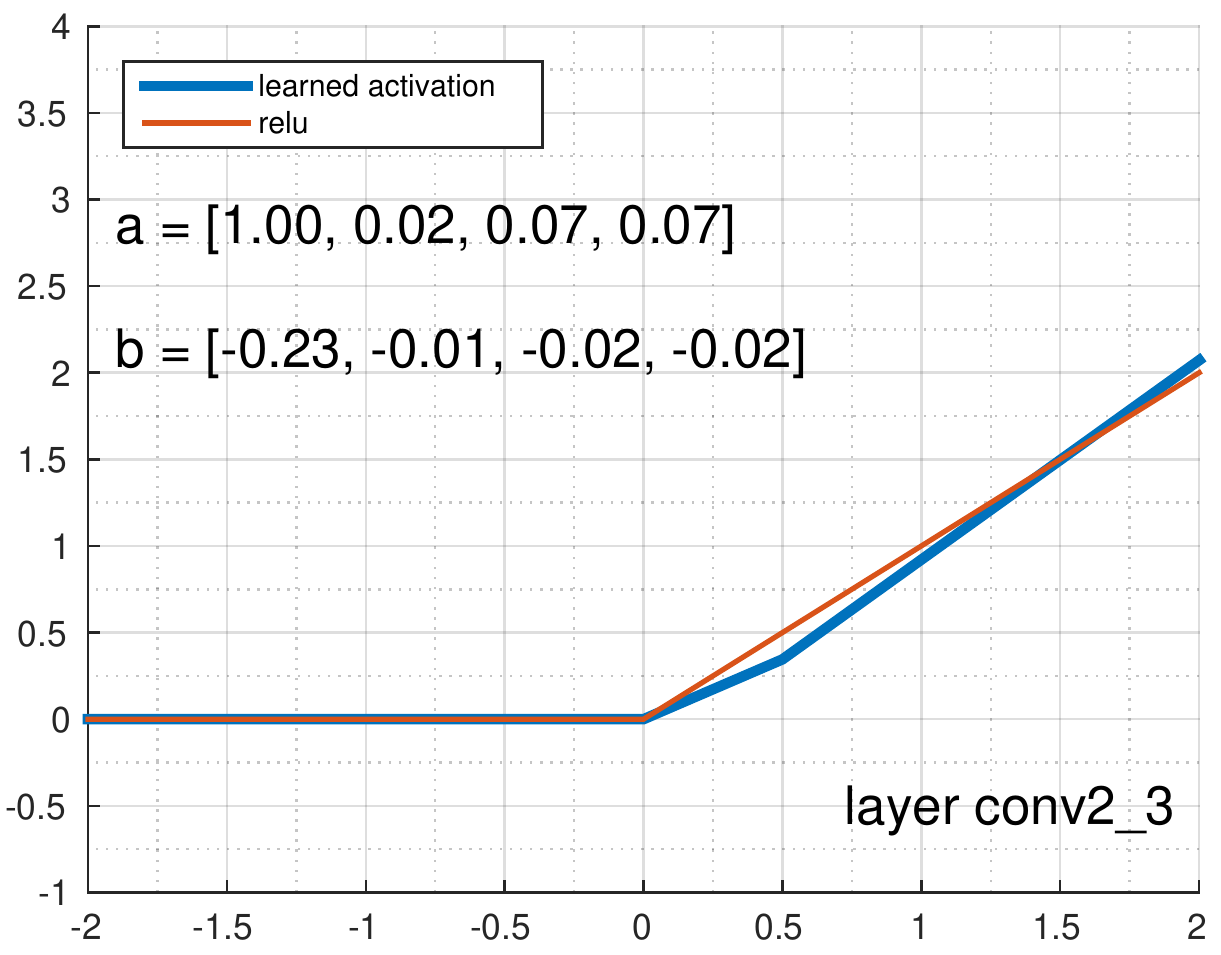} \\
		(a) ReLU-shape
	\end{minipage}
	\begin{minipage}[b]{0.24\linewidth}
		\centering
		\includegraphics[width=1\linewidth]{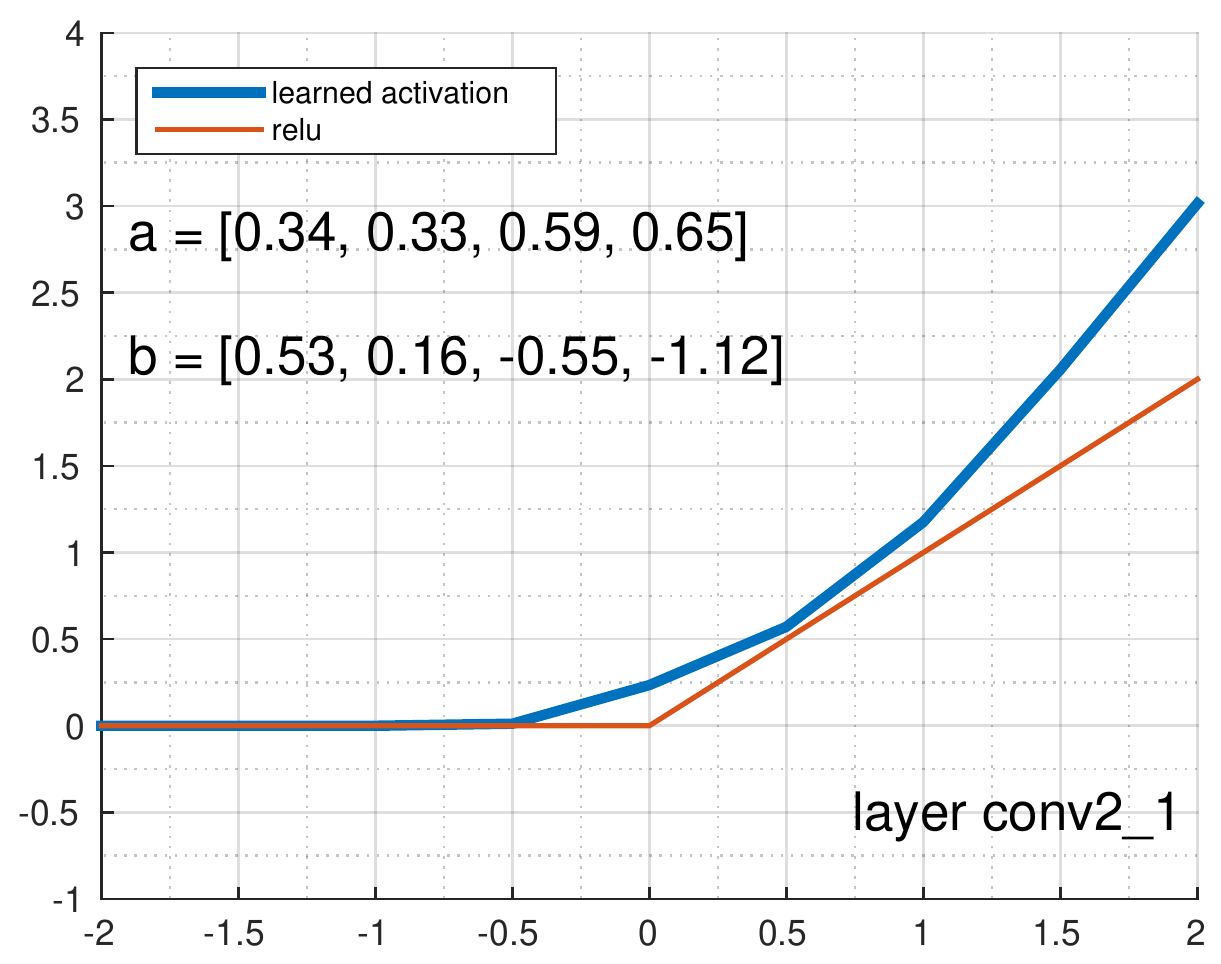} \\
		(b) Shifted exponential-shape
	\end{minipage}
	%\\
	\begin{minipage}[b]{0.24\linewidth}
		\centering
		\includegraphics[width=1\linewidth]{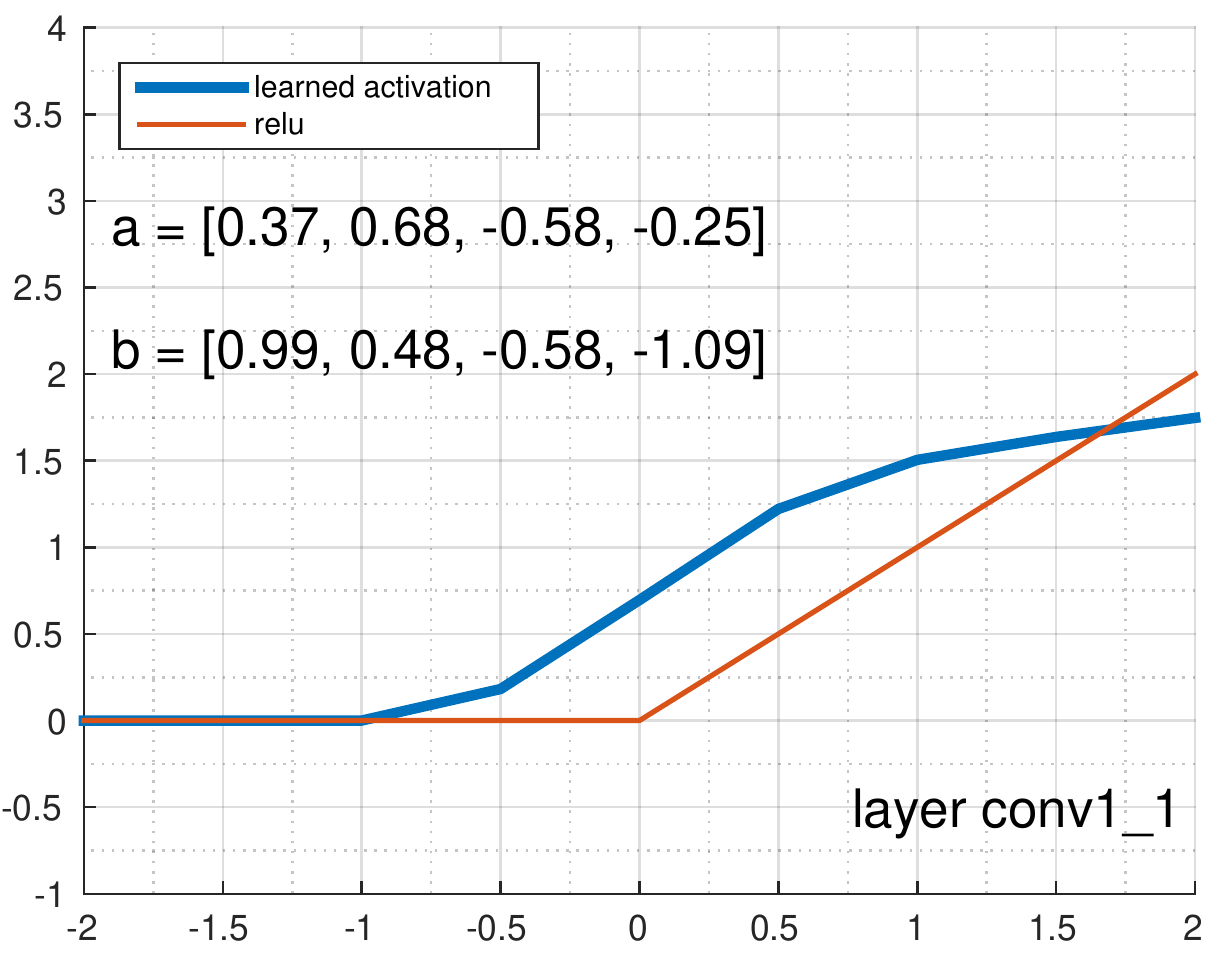} \\
		(c) Sigmoid-shape
	\end{minipage}
	\begin{minipage}[b]{0.24\linewidth}
		\centering
		\includegraphics[width=1\linewidth]{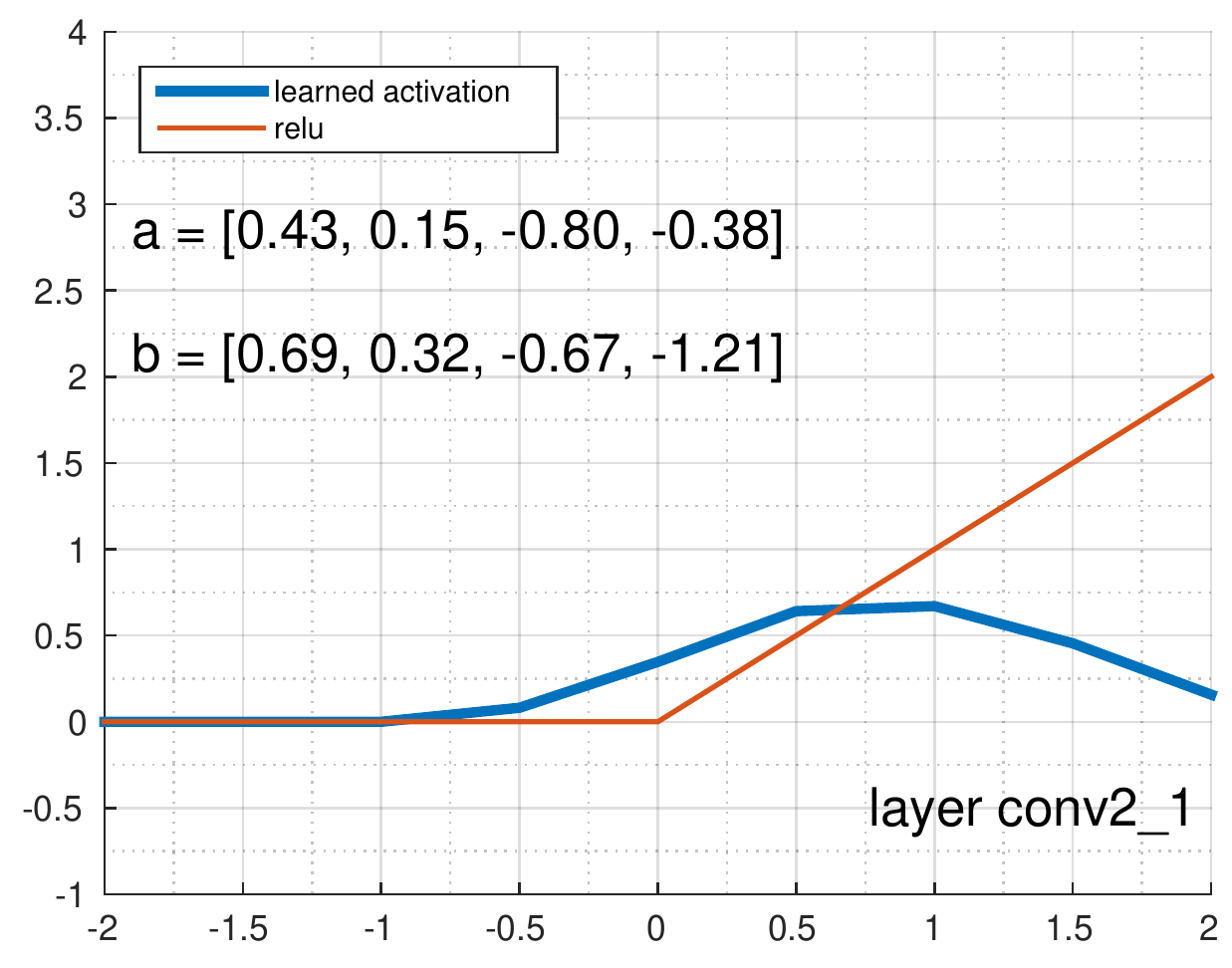} \\
		(d) Trapezoid-shape
	\end{minipage}
	\\
	\\
	\begin{minipage}[b]{0.24\linewidth}
		\centering
		\includegraphics[width=1\linewidth]{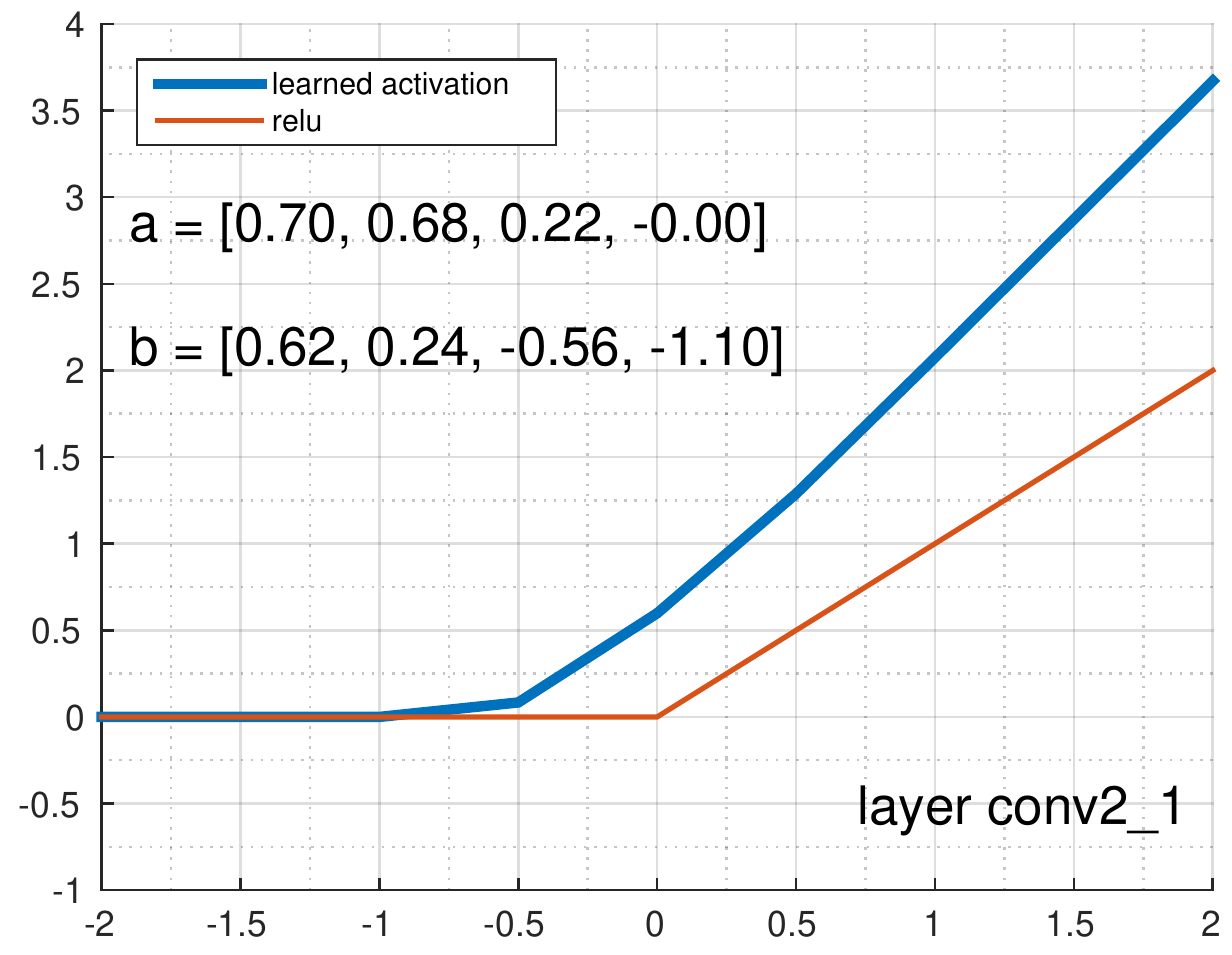} \\
	\end{minipage}
	\begin{minipage}[b]{0.24\linewidth}
		\centering
		\includegraphics[width=1\linewidth]{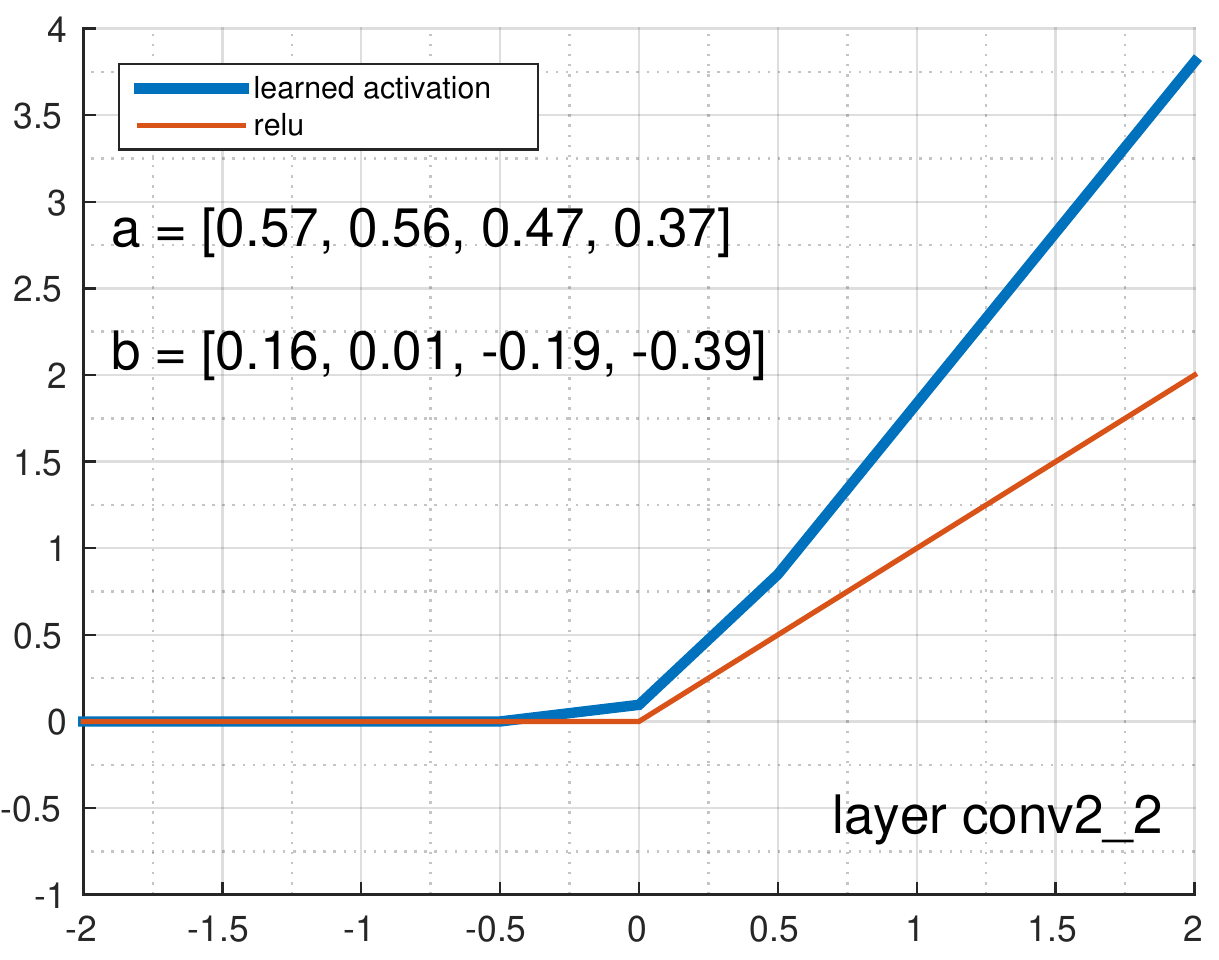} \\
	\end{minipage}
	\begin{minipage}[b]{0.24\linewidth}
		\centering
		\includegraphics[width=1\linewidth]{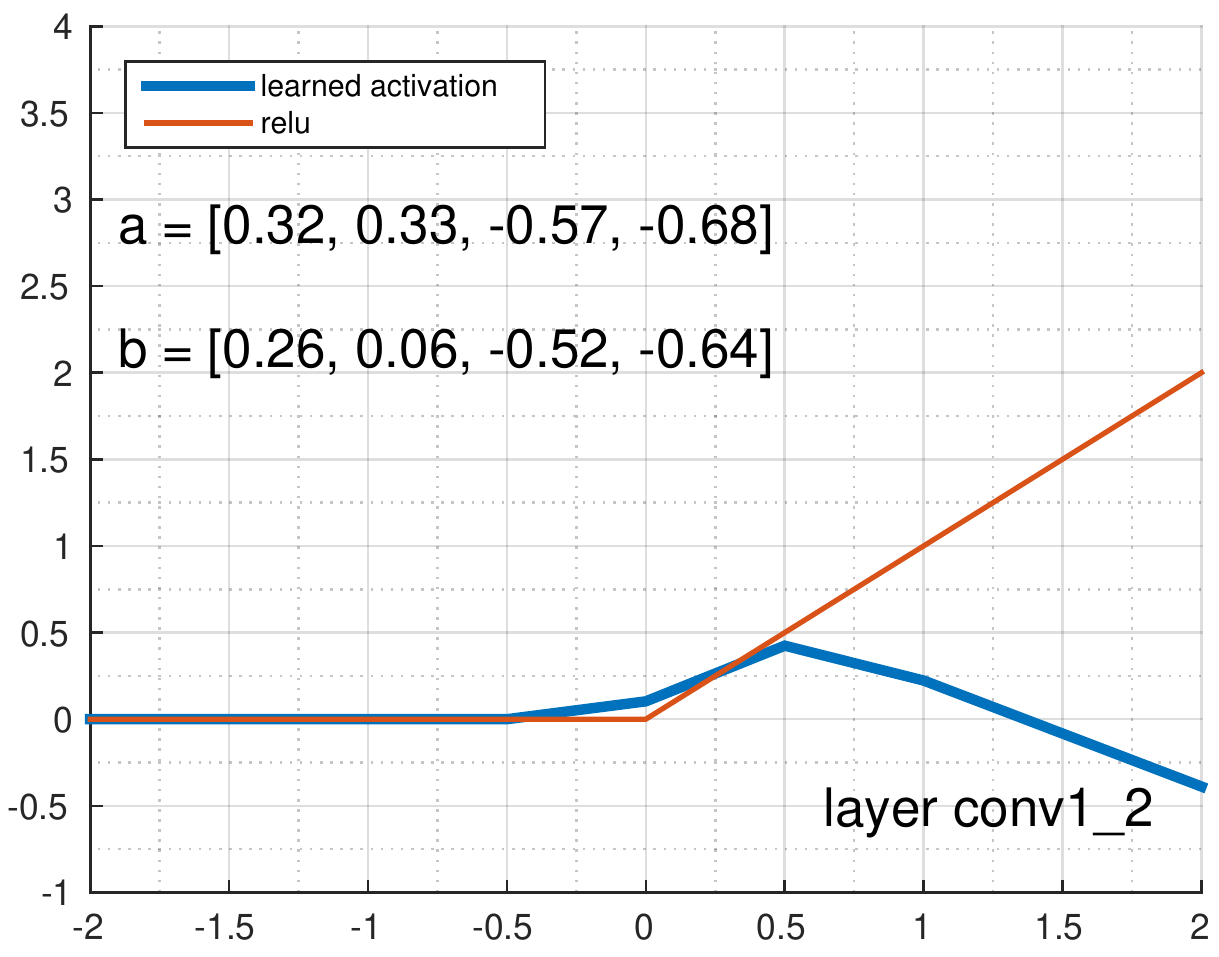} \\
	\end{minipage}
	\begin{minipage}[b]{0.24\linewidth}
		\centering
		\includegraphics[width=1\linewidth]{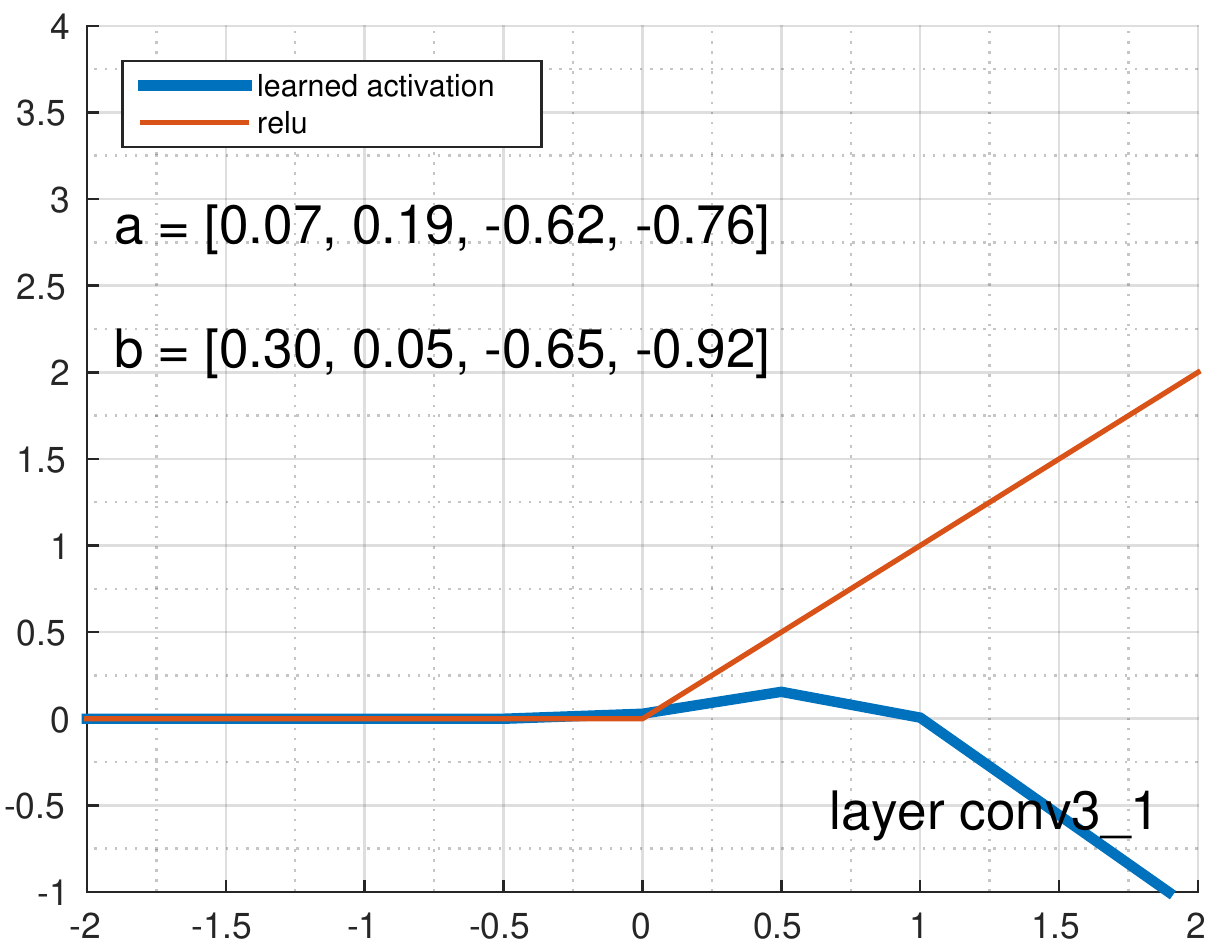} \\
	\end{minipage}
	\\
	\begin{minipage}[b]{0.24\linewidth}
		\centering
		\includegraphics[width=1\linewidth]{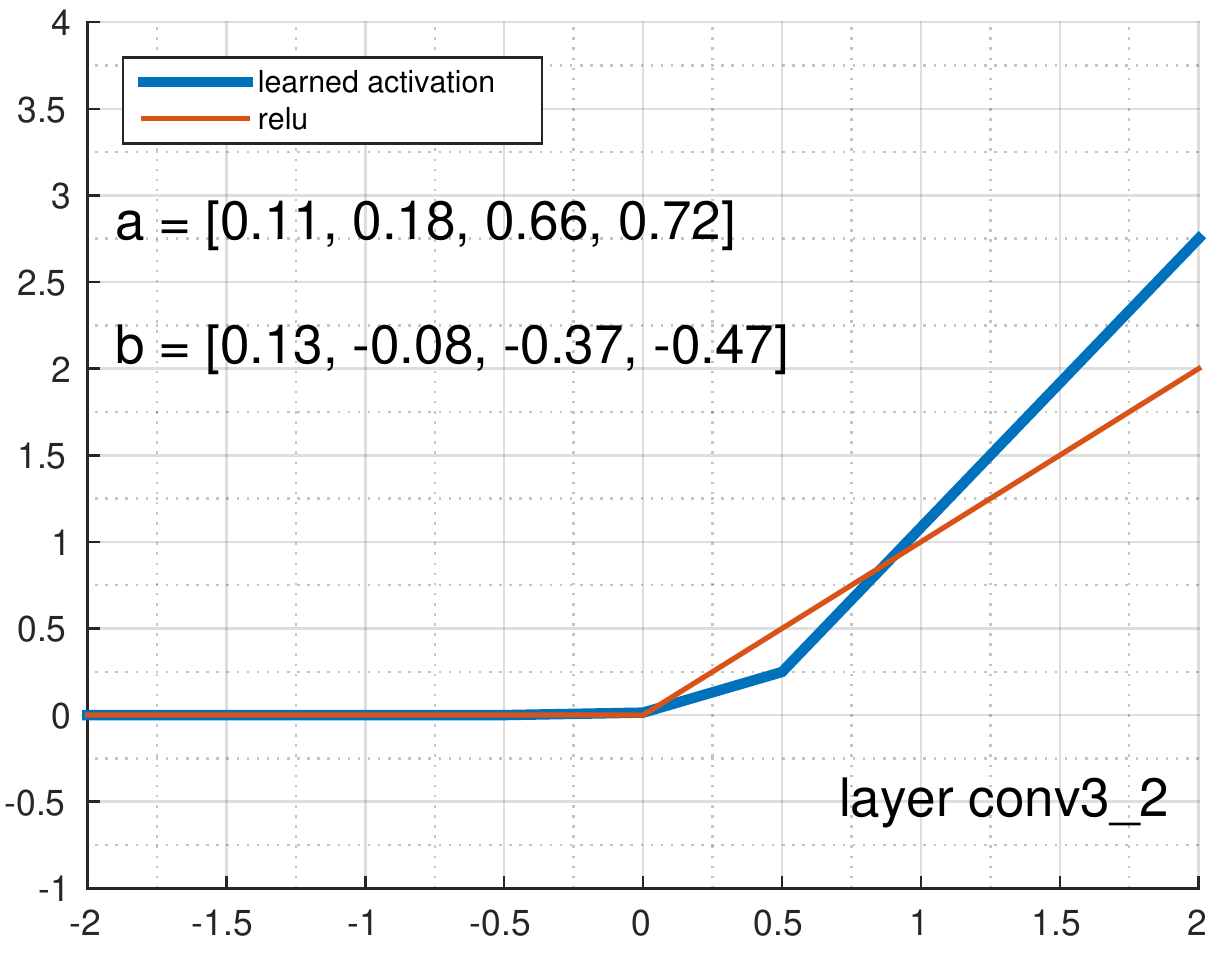} \\
	\end{minipage}
	\begin{minipage}[b]{0.24\linewidth}
		\centering
		\includegraphics[width=1\linewidth]{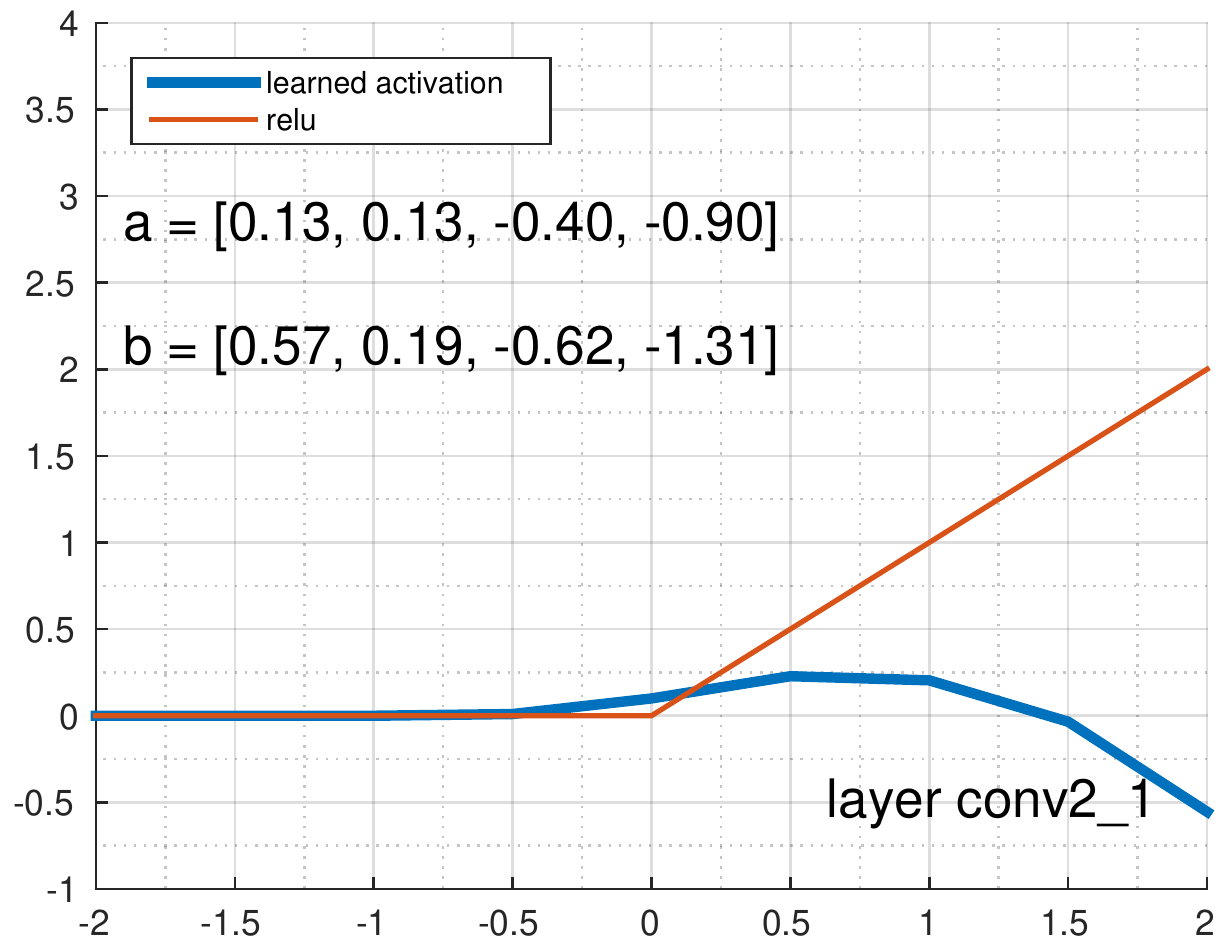} \\
	\end{minipage}
	\begin{minipage}[b]{0.24\linewidth}
		\centering
		\includegraphics[width=1\linewidth]{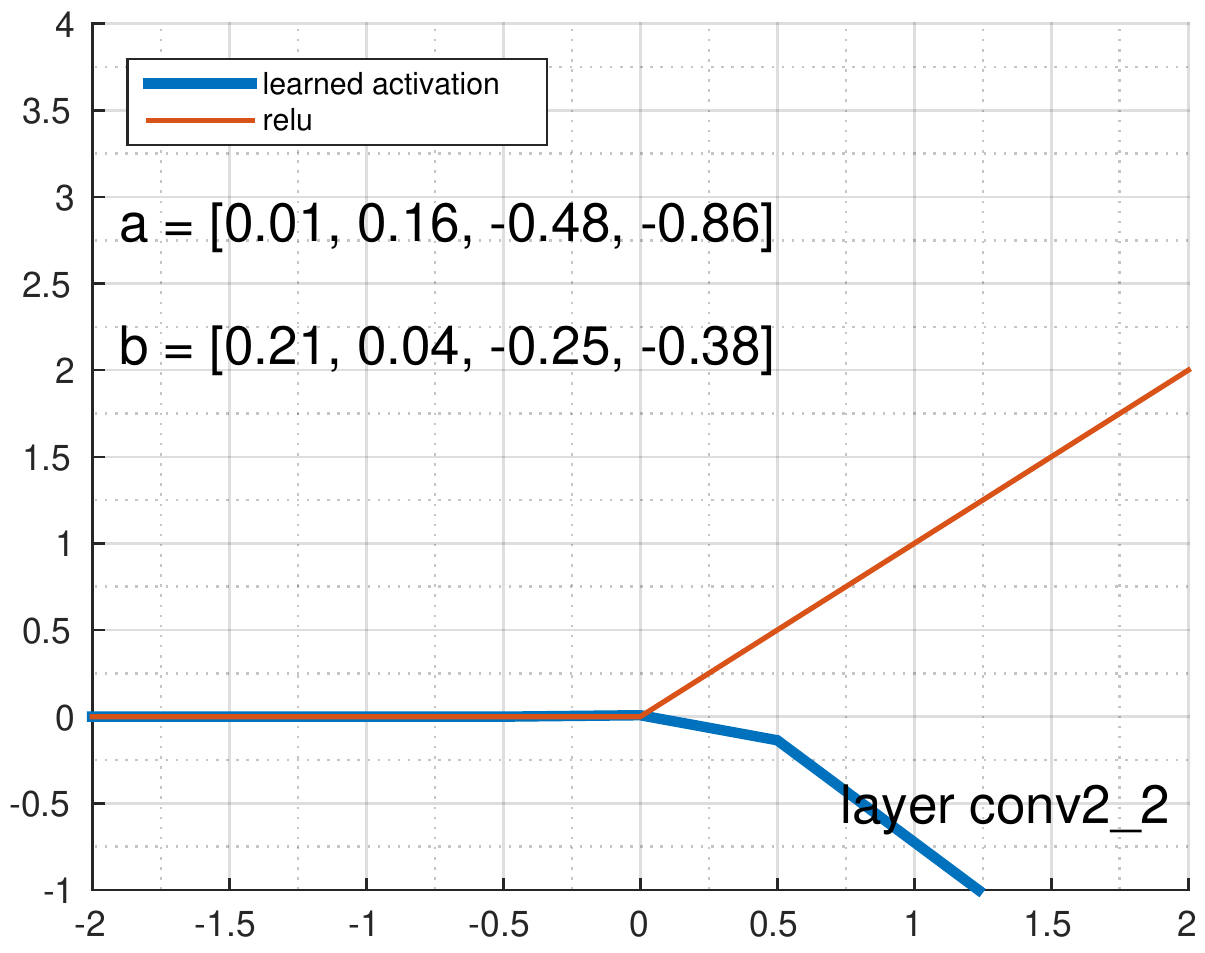} \\
	\end{minipage}
	\begin{minipage}[b]{0.24\linewidth}
		\centering
		\includegraphics[width=1\linewidth]{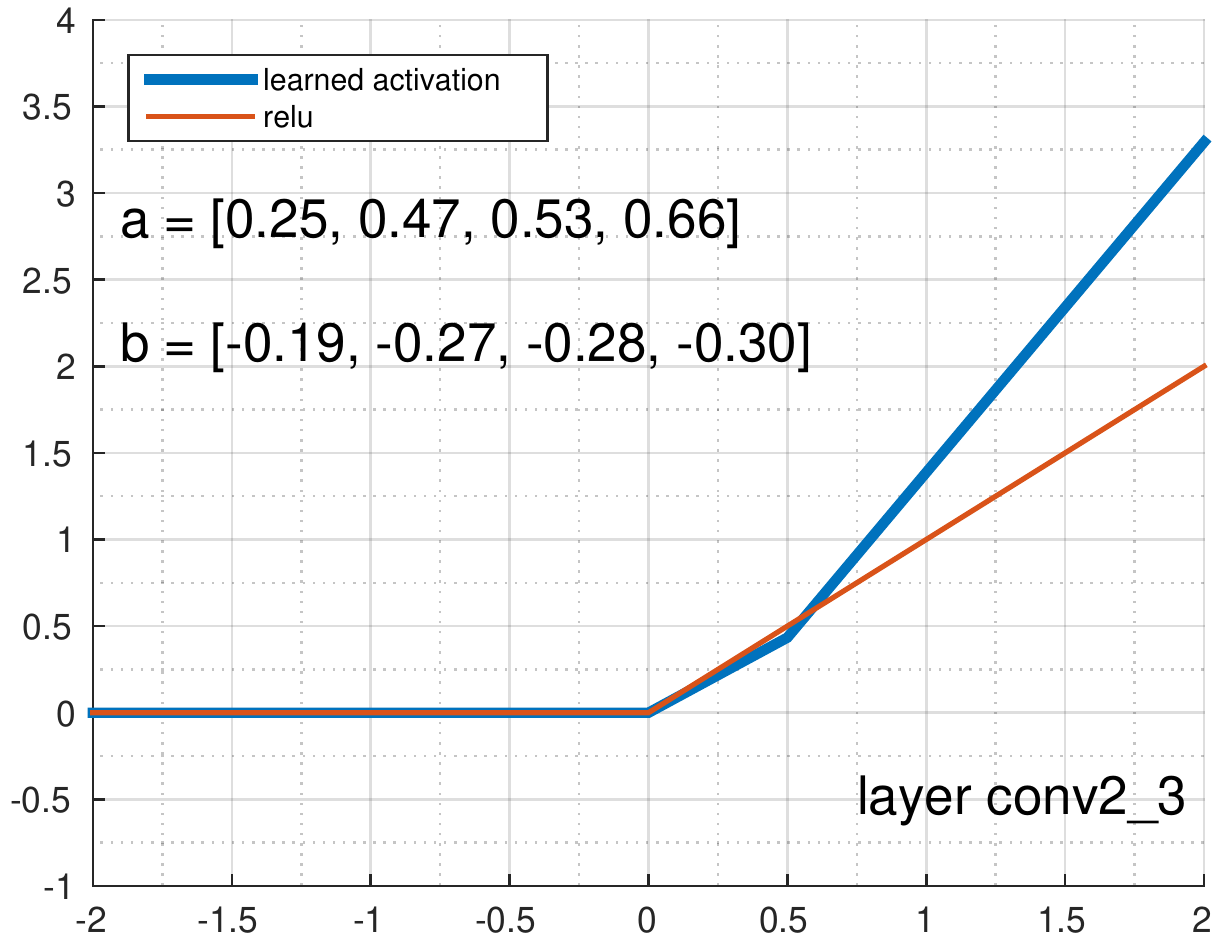} \\
	\end{minipage}
	\vspace{-.1cm}
	\caption{The learned MBA parameters $a, b$ in a certain layer ($K=4$). We use the MBA model $\#9$ in Table \ref{tab:ablation} as an illustration. 
		Each figure plots the mapping function $B_{m,n,j,i}$. The horizontal axis indicates $x_{n,i}$ in the input feature map and vertical axis indicates $u_{m,n,j,i}$ in the output map in
		(\ref{eqn:bias3}). The index of feature channels $m$ and $n$, and locations $j$ and $i$ are dropped to be concise. 
		The fist row shows four typical shape during the activation learning whilst the rest gives more visual examples.}\label{fig:learned_param}
\end{figure*}

\subsection{MBA with its subsequent convolutional layer}
The output of the MBA module is a set of maps $\{\hat{\mathbf{x}}_{n,k}|n=1, \ldots, N, k=1, \ldots, K\}$. These maps are then linearly combined 
by its subsequent convolutional layer into $\mathbf{h}_m$ as follows:
\begin{align}\label{eqn:bias1}
\begin{split}
  \mathbf{h}_m = \sum_{n=1}^{N} \sum_{k=1}^{K} \mathbf{W}_{m,n,k} \hat{\mathbf{x}}_{n,k} \\
  \end{split}
\end{align}
where $m=1, \ldots, M$ and $\mathbf{W}_{m,n,k} \hat{\mathbf{x}}_{n,k}$ is the representation of convolution by matrix multiplication.
Denote the $j$-th element in $\mathbf{h}_m$ by $h_{m,j}$,  we have
\begin{align}\label{eqn:bias2}
\begin{split}
  h_{m,j} =   \sum_{n=1}^{N} \sum_{k=1}^{K} \sum_{i=1}^{WH} w_{m,n,k,i,j}  \hat{{x}}_{n,k,i}, \\
  \end{split}
\end{align}
where $w_{m,n,k,i,j}$ denotes the $(j, i)$-th element in the matrix $\mathbf{W}_{m,n,k}$. 
Taking the representation of $\hat{x}$ by $x$ in (\ref{eqn:bias_ele2}) into consideration, we have the factorized version of (\ref{eqn:bias2}):
\begin{align}\label{eqn:bias3}
\begin{split}
  h_{m,j} &=  \sum_{i=1}^{WH} \sum_{n=1}^{N} \sum_{k=1}^{K} w_{m,n,k,j,i} \sigma( x_{n,i} +b_{n,k}), \\
&=\sum_{i=1}^{WH}  \sum_{n=1}^{N} w_{m,n,j,i}' \sum_{k=1}^{K} a_{m,n,k,j,i} \sigma(x_{n,i} +b_{n,k}), \\
&=\sum_{i=1}^{WH}  \sum_{n=1}^{N} w_{m,n,j,i}' u_{m,n,j,i}, \\
\end{split}
\end{align}
where $w'$ and $u$ take the forms of:
\begin{align}
 w_{m,n,k,j,i} \stackrel{\vartriangle}{=} &~w_{m,n,j,i}'  a_{m,n,k,j,i}, \\
u_{m,n,j,i} \stackrel{\vartriangle}{=} &~\sum_{k=1}^{K} a_{m,n,k,j,i} \sigma(x_{n,i} +b_{n,k}), \label{eqn:bias4}
\end{align}
here $w'$ is an intermediate variable to generate the coefficient $a_{m,n,k,j,i}$.
%%%description%%%%%
The formulation in (\ref{eqn:bias4}) shows that the element $x_{n,i}$ in the feature map is separated by multiple biases to obtain multiple ReLU functions $\sigma(x_{n,i}+b_{n,k})$, and then these ReLU functions are linearly combined by the weights $a_{m,n,k,j,i}$ to obtain $u_{m,n,j,i}$, which serves as the decoupled pattern in the MBA layer for the $m$-th channel in the next convolutional layer at location $j$.
The $j$-th element in $\textbf{h}$, \textit{i.e.}, $h_{m,j}$ is a weighted sum of $u_{m,n,j,i}$ as shown in (\ref{eqn:bias3}). Therefore, the key is to study the mapping $x_{n,i}$ at location $i$ in an input feature map to $u_{m,n,j,i}$ at location $j$ in an output feature map in (\ref{eqn:bias4}). Such a mapping is \textit{across} feature channels and locations.  (\ref{eqn:bias4}) can be defined as a mapping function $u_{m,n,j,i} = B_{m,n,j,i}(x_{n,i})$. There is a large set of such mapping functions $\{B_{m,n,j,i}\}$ which are characterized by parameters $\{a_{m,n,k,j,i}\}$ and $\{b_{n,k}\}$.
In the following discussion, we skip the subscripts $_{m,n,j,i}$ to be concise.

We show the learned
parameters of $a_{k}$ and $b_{k}$ for the input $x$ and the decoupled pattern $u$ in Figure \ref{fig:learned_param}.
%To investigate how the bias layer takes effect, we show the learned parameters of $a$ and $b$ in .
%Fig.\ref{fig:learned_param} depicts some patterns captured by the learned $g(x)$ in certain layers.
%
%We can find that these observations are in conjunction with Thm.\ref{pythagorean} aforementioned.
%
Specifically, Fig.\ref{fig:learned_param} (a) approximates the ReLU unit where $a\approx [1,0,0,0]$ is the base along the first axis in a 4-dimension space;
Fig.\ref{fig:learned_param} (b) displays the property of leaky ReLU \cite{leacky_relu} to allow gradient back-propagation around small negative input;
moreover, it has a steeper slope in the positive region than ReLU, which makes the training process faster.
%
%If we change to the value of $a_k$ around zero input to be negative, a `pull-down' version of (b) can result in Fig.\ref{fig:bias}(c), where $[a_2, b_2] = [-0.47, 0.08]$
%and the activation is locally concave.
%
Fig.\ref{fig:learned_param} (c) stimulates the case where the activation function serves as a sigmoid non-linearity, \textit{i.e.},
only allowing small values around zero to backpropagate the gradients.
The non-linear activation function in Fig.\ref{fig:learned_param} (d) forms a trapezoid-shaped and serves as the histogram bin `collector' - only selecting the input $x$
within a small range and shuttering all the other activations outside the range.
Note that the mapping is concave
%also shares similar mechanism of leaky ReLU and local `pull-up' as in (b) and (c),
%apart from it is almost globally concave
due to the negative values of $a_{ik}$ when %the input goes to positive infinity.
$x\rightarrow + \infty$, in which case neither the standard ReLU nor APL unit could describe.
In addition, the second and third rows of Figure \ref{fig:learned_param} show more examples of the mappings decomposed from parameters in the convolution layer,
from which we can see a wide diversity of patterns captured by the multi-bias mechanism to decouple the signal strength
and the cross-channel sharing scheme to combine visual representations in a much more flexible way.

\begin{figure}[t]
	\centering
	\includegraphics[width=.8\linewidth]{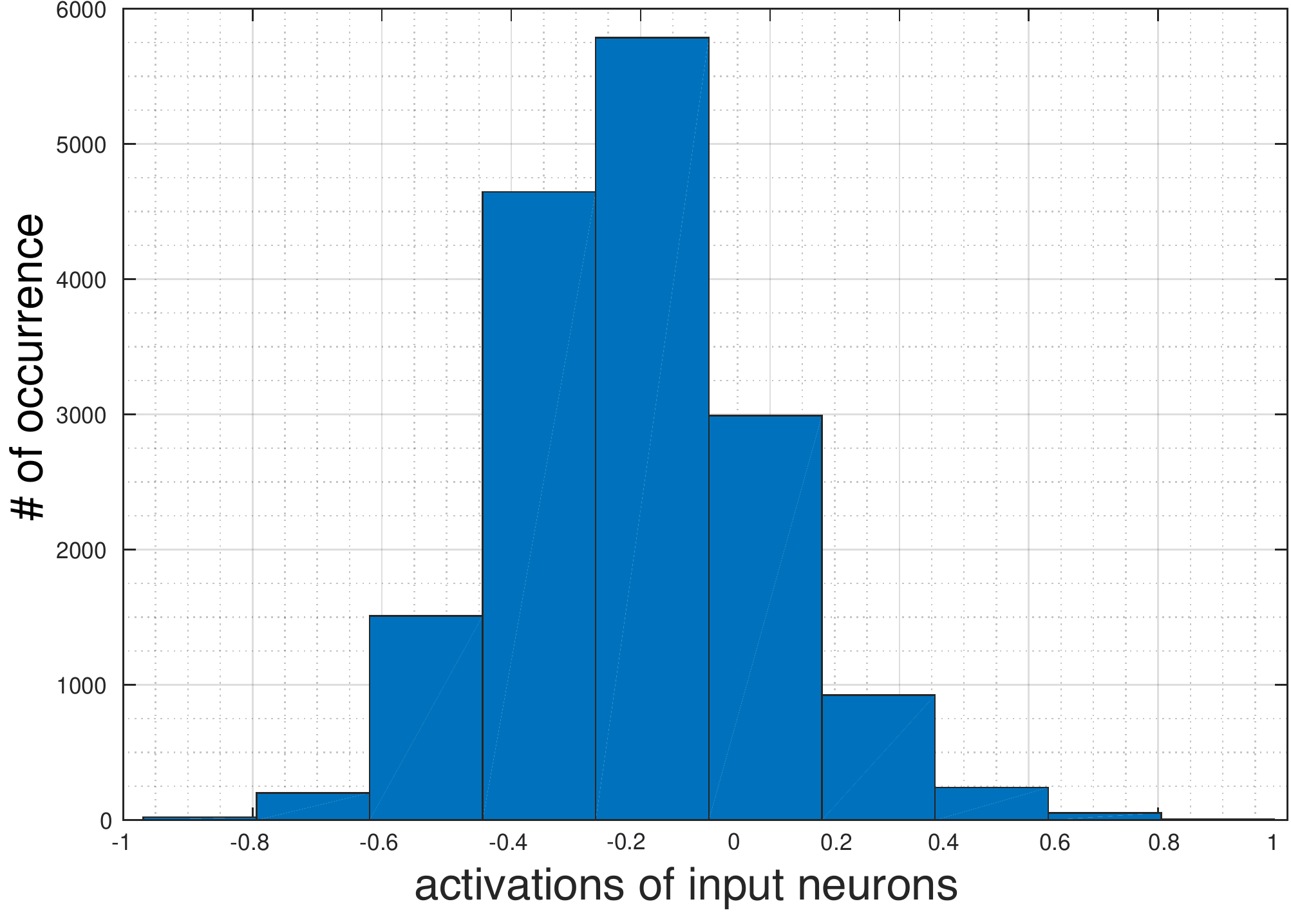}\\
	\vspace{-.15cm}
	\caption{Histogram of the neuron activations before the MBA module, of which most activations are sparse
		and centered around zeros.}\label{fig:hist}
	\vspace{-.3cm}
\end{figure}

Figure \ref{fig:hist} shows the histogram of response magnitudes in the input feature map before feeding into the MBA module.
We adopt the architecture of model $\#6$ in Table \ref{tab:ablation} on CIFAR-10 to obtain the distribution result,
where we randomly select 1,000 samples to get the average activations over all the MBA output.
%
%We can see that the input distribution is sparse and focused around zero,
%which in turn explains the functionality of the activation function in Fig.\ref{fig:learned_param}(d).
The histogram distributes around zero, which indicates the learned pattern $u$ or mapping $B$ affects the neurons only in a small range near zero.

%\begin{figure}
%  \centering
%  \includegraphics[width=\linewidth]{piecewise}\\
%  \vspace{-.15cm}
%  \caption{The pipeline of the piecewise linear function. Note that such a scheme does not share visual patterns
%  of different feature maps after the APL unit and thus leads to inferior expressive power in the feature space.}\label{fig:piecewise}
%\end{figure}

\subsection{Relationship with piecewise-linear functions}\label{comparison}
To see further the advantage of our algorithm, we compare our MBA layer with the recently proposed method called adaptive piecewise linear function unit (APL) \cite{piecewise_linear}.
It formulates the activation function as a sum of hinge-shaped functions.
Fig.\ref{fig:bias} (b) describes the pipeline of the APL module.
Given the feature map $\mathbf{x}_n$, it
generates the output $\hat{x}_{n,k,i}$ from $k$ piecewise linear functions from each element $x_{n,i}$ in $\mathbf{x}_n$ as follows:
\begin{align}\label{eqn:piecewise}
\hat{x}_{n,i}= \max(0,x_{n,i}) + \sum_{k=1}^{K} a_{k,i} \max(0, -x_{n,i}+b_{k,i}),
\end{align}
where $ a_{k,i}, b_{k,i}$ are learnable parameters to control the shape of the non-linearity function.
Then the subsequent convolutional layer computes the weighted sum of `piece-wise-linearized' maps to have the output $h_{m,j}$ of channel $m$ at location $j$ as follows:
\begin{align}\label{eqn:bias5}
\begin{split}
  h_{m,j} &=  \sum_{i=1}^{WH} \sum_{n=1}^{N} w_{m,n,i,j}  \left(\sum_{k} { a_{k,i}\max(0, -x_{n,i}+b_{k,i})}\right),\\
  &=  \sum_{i=1}^{WH} \sum_{n=1}^{N} w_{m,n,i,j} u_{n,i},\\
  \end{split}
\end{align}
where $w$ is the parameters of the subsequent convolution kernel and we define 
\begin{align}
u_{n,i} \stackrel{\vartriangle}{=} &~\sum_{k=1}^{K} a_{k,i} \max(0, -x_{n,i}+b_{k,i}), \label{eqn:bias_shit}
\end{align}
in a similar derivation through (\ref{eqn:bias1}) to  (\ref{eqn:bias3}).
It is obviously seen that APL represented in (\ref{eqn:bias5}) is a special case of MBA by enforcing $u_{m_1,n,j,i} = u_{m_2,n,j,i}, \forall m_1, m_2$ in (\ref{eqn:bias4}). Therefore, for different target channel with index $m_1$ and  $m_2$,
 the piecewise-linear function provides the \textit{same} $u_{*}$ while MBA provides \textit{different} $u_{*}$. Take again the case in Figure \ref{example} for instance, the output channel $m_1=1$ for eyes requires $u$ with high magnitude while $m_2=2$ for mouth requires $u$ with low magnitude. 
 This kind of requirement cannot be met by APL but can be met by our MBA. 
 This is because our MBA can separate single $x$ into different $u_{*}$ for different $m$ according to the magnitude of $x$ while APL cannot.

A closer look at the difference between our algorithm and the APL unit is through the two diagrams in Figure \ref{fig:bias}. When an input feature map
is decoupled into multiple band maps after the biasing process, APL only recombines band maps from the same input feature map to generate one output.
However, MBA concatenates band maps across different input maps and allows to select and combine them in a flexible way, thus generating a much richer set of maps (patterns) than does the APL.

\renewcommand{\arraystretch}{1.2}
\begin{table}[pt]
	\centering
	\caption{Investigation of the MBA module in different architectures on the MNIST and CIFAR-10 dataset. 
	We choose $K=4$ for both MBA and APL.
    %The architecture of the shallow and deep networks are defined in the context.
    % contains three convolution layers while the deep one consists of nine layers. 
    Bracket [$\cdot$] denotes a stack of three convolution layers. %with the same number of output maps.
    `VaCon' means the vanilla neural network. `@' represents a different number of output maps before the MBA module, see context for details.}\label{tab:cmp_model}
	\medskip
    \footnotesize{
	\begin{tabular}{l|l|l|l}
		\hline
		\multicolumn{1}{l|}{Model}      & \# Params  & MNIST & CIFAR-10 \\ \hline
        \multicolumn{4}{c}{\emph{Shallow network: 32-64-128-1024-1024}} \\ \hline\hline
		\multicolumn{1}{l|}{VaCon}            &     93k  & 2.04\%      & 34.27\%\\ \hline
		\multicolumn{1}{l|}{VaCon@4x}            &  594k     & 0.95\%      & 22.75\%\\ \hline
		\multicolumn{1}{l|}{APL }        &    120k   & 1.08\%      & 28.72\%\\ \hline
		\multicolumn{1}{l|}{APL@4x }        &   620k    & 1.15\%      & 23.80\%\\ \hline
		\multicolumn{1}{l|}{APL@same }      &   358k    & 1.17\%      & 31.53\%\\ \hline
		\multicolumn{1}{l|}{MBA}                   &    369k   & 0.83\%      & 22.39\% \\ \hline
        \multicolumn{4}{c}{\textit{Deep network:} [\textit{32}]-[\textit{64}]-[\textit{128}]-\textit{1024}-\textit{1024}} \\ \hline\hline
        \multicolumn{1}{l|}{VaCon}            &   480k    & 1.03\%            & 19.25\%\\ \hline
        \multicolumn{1}{l|}{VaCon@4x}            &  7.6M     & 0.42\%      & 14.38\%\\ \hline
		\multicolumn{1}{l|}{APL }        & 550k     & 1.37\%           & 22.54\%\\ \hline
		\multicolumn{1}{l|}{APL@4x }        & 7.7M      & 0.54\%      & 13.77\%\\ \hline
		\multicolumn{1}{l|}{APL@same }        & 2.1M     & 0.82\%      & 17.29\%\\ \hline
		\multicolumn{1}{l|}{MBA}                   & 2M      & 0.31\%          & 12.32\% \\ \hline
	\end{tabular}
}
\vspace{-.2cm}
\end{table}

Table \ref{tab:cmp_model} shows the investigation breakdown on MNIST and CIFAR-10 when comparing MBA with
APL
and vanilla neural networks\footnote{ We do not use the tricks presented in APL \cite{piecewise_linear}, for example, changing the dropout ratio or adding them before certain pooling layer.}.
Here we adopt two types of network: the shallow one has three convolutional layers with the number of output 32, 64, 128, respectively
and two fully connected layers which both have an output neurons of 1024;
the deep one has nine convolutional layers divided into three stacks with each stack having the same number of output 32, 64, 128 and two fully connected layers.
All convolutional layers have the same kernel size of 3 and we use max pooling of size 2, stride 2, at the end of each convolution or stack layer.
Also we keep training parameters the same across models within each architecture to exclude external factors.
%,
%including momentum (0.9), learning rate (0.1) and its policy (step), weight decay (0.0005) and dropout ratio (0.5).

%COMPUTATION
The number of parameters in each setting of Table \ref{tab:cmp_model} does not count those in the fully connected layers
and we can compare the computational cost of MBA, APL and vanilla net quantitatively. Take the deep architecture case for example,  
the vanilla network has about $480$k parameters with the designated structure; by applying the MBA module on it, the additional parameters are 
(a) the increasing channels of kernels
in each subsequent layer, \textit{i.e.}, $Nq^2M(K-1)$, where $q=3$ is the kernel size; and
(b) a small fraction of the bias term, $NK$. Therefore we have a computational cost of $480\text{k}+1.5\text{M}+3\text{k}=2\text{M}$. 
However, if we force the vanilla model to have the same output maps to be fed into the subsequent layers (models denoted as `@4x'), there has to be $N(K-1)$
more maps coming from the convolutional kernel in the current layer. As mentioned in Section \ref{intro}, such a scheme would increase
the parameter overhead of kernels to a great extent. That is approximately $K$ times the size of the MBA module (in this case, $7.6$M vs $2$M).

%ANALYSIS
Several remarks can be drawn from Table \ref{tab:cmp_model}.
%\vspace{-.2cm}
%\begin{itemize}
%	\item 
First, both MBA and APL modules imposed on the vanilla net can reduce test errors. 
Second,
as the number of feature maps increases, vanilla networks can further boost the performance. 
However, it is less inferior compared with the MBA module
(0.42\% vs 0.31\% on MNIST) where the latter has a much smaller set of parameters.
Third,
the piecewise linear function does not perform well compared with the proposed method, even though it has the same network width (APL@4x)
	or similar parameters (APL@same, by changing the output number of feature maps) as in the MBA model.
This is probably due to the limited expressive power, or inferior ability of feature representation in (\ref{eqn:bias_shit}). 
%\end{itemize}
%
%Compared with the piecewise linear function, our model further reduces the error by 6.4\% and 1.5\% on 
%CIFAR-10 for shallow and deep architecture respectively.
Therefore, these observations  further proves
the importance of applying the MBA module to separate the responses of various signals and feed the across-channel 
information into the next layer in a simple way, instead of 
buying more convolutional kernels.

%\begin{figure}
%  \centering
%  \includegraphics[width=\linewidth]{piecewise}\\
%  \vspace{-.15cm}
%  \caption{The pipeline of the piecewise linear function. Note that such a scheme does not share visual patterns
%  of different feature maps after the APL unit and thus leads to inferior expressive power in the feature space.}\label{fig:piecewise}
%\end{figure}

%%%%%%%%%%%%%%%%%%%%%%%%%%%%%%%%%%%%%%%%%%%%%%%%%%%%%%%%%%%%
%%%%%%%%%%%%%%%%%%%%%%%%%%%%%%%%%%%%%%%%%%%%%%%%%%%%%%%%%%%%%%%%%%
\section{Experimental  results}\label{exp}

We evaluate the proposed MBA module and compare with other state-of-the-arts on 
several benchmarks.
The CIFAR-10 dataset \cite{cifar} consists of $32 \times 32$ color images on 10 classes and is divided into 50,000 training images and 10,000 testing images.
The CIFAR-100 dataset has the same size and format as CIFAR-10, but contains 100 classes, with only one tenth as many labeled examples per class.
%STL-10 \cite{stl} is designed to test unsupervised learning algorithms and hence has a relatively small labeled training set of 500 images per class.
%The test set contains 800 labeled images per class. All examples are $96\times96$ pixel color images acquired from ImageNet.
The SVHN \cite{svhn} dataset resembles MNIST and consists of color images of house numbers captured by Google street view.
We use the seoncd format of the dataset where each image is of size $32\times 32$ and the task is to classify the digit in the center. 
Additional digits may appear beside it and must be ignored.
All images are preprocessed by subtracting each pixel value by the mean computed from the corresponding training set. We follow a similar split-up of the validation set
from the training set as \cite{maxout}, where one tenth of samples per class from the training set on CIFAR, and 
400 plus 200 samples per class from the training and the extra set on SVHN, are selected to build a validation set.%, respectively.

\subsection{Implementation Details}
\label{sec:imple_detail}

Our baseline network has three stacks of convolutional layers with each stack containing three convolutional layers, 
resulting in a total number of nine layers.
Each stack has [96-96-96], [128-128-128] and [256-256-512] filters, respectively.
The kernel size is $3$ and padded by $1$ pixel on each side with stride $1$ for all convolutional layers.
At the end of each convolutional stack is a max-pooling operation with kernel and stride size of 2.
The two fully connected layers have 2048 neurons each. We also apply dropout %\cite{dropout} 
with ratio $0.5$ after each fully connected layers.
The final layer is a softmax classification layer. 
%We use the Rectified linear units as the non-linearity function throughout the baseline and all variations of MBA models.

The optimal training hyperparameters are determined on each validation set.
We set the momentum as 0.9 and the weight decay to be 0.005. 
The base learning rate is set to be $0.1, 0.1, 0.05$, respectively.
We drop the learning rate by 10\% around every 40 epoches in a continuous exponential way and stop to decrease the learning rate until it reaches
a minimum value ($0.0001$). For the CIFAR-100 and SVHN datasets, we use a slightly longer cycle of 50 epoches to drop the rate by 10\%.
The whole training process takes around 100, 150, and 150 epoches on three benchmarks.
We use the hyperparameter $K=4$ for the MBA module  %i.e. each channel is decoupled into 4 channels in the MBA layer. 
and the mini-batch size of $100$ for stochastic gradient descent.
All the convolutional layers %in the deep model 
are initialized with Gaussian distribution with mean of zero and standard variation of $0.05$ or $0.1$.
%
%As for the initialization of the bias layer, 
We do not carefully cross-validate the initialized biases in the MBA module to find the optimal settings
but simply choose a set of constants to differentiate the initial biases.

\subsection{Ablation Study}
\renewcommand{\arraystretch}{1.5}
\begin{table}[t]
	\centering
	\caption{Ablation study of applying the bias module with different width and depth into the network on CIFAR-10.
		Empty entry means that MBA is not included % and the number of filters is the same as the baseline  in the corresponding stack. 
		while the rest employs a MBA module and specifies the number of output feature maps for the corresponding convolution layer. 
		%Numbers indicate the number of output feature maps for the convolution layers in the stack.
		}\label{tab:ablation}
	\medskip
	\scriptsize{
		%\footnotesize{
		\begin{tabular}{l|l|l|l|l}
			\hline
			Model      & Conv-1 & Conv-2 & Conv-3 & Test Error. \\ \hline
			Baseline         & 96-96-96  & 128-128-128 & 256-256-512 & 9.4\%   \\ \hline\hline
			1 &\multicolumn{1}{c|}{-}   &\multicolumn{1}{c|}{-}  &64-64-128& 8.5\% \\
			2       & \multicolumn{1}{c|}{-}&\multicolumn{1}{c|}{-}&128-128-256& 7.3\%    \\
			3       & \multicolumn{1}{c|}{-} &\multicolumn{1}{c|}{-}&256-256-512& 7.2\%  \\ \hline
			4       & \multicolumn{1}{c|}{-} &32-32-32&64-64-128& 10.4\%  \\
			5 &\multicolumn{1}{c|}{-} &64-64-64 & 128-128-256& 8.2\%   \\
			6       & \multicolumn{1}{c|}{-} &128-128-128 & 256-256-512& \textbf{6.7\%}   \\ \hline
			7       &  24-24-24 & 32-32-32 & 64-64-128      & 11.7\%   \\
			8       &  48-48-48&64-64-64 & 128-128-256   & 8.8\%   \\
			9       &  96-96-96&128-128-128 & 256-256-512     & 6.8\%   \\ \hline
		\end{tabular}
	}
\end{table}

\renewcommand{\arraystretch}{1.2}
\begin{table}
	\centering
	\caption{Effects of the hyperparameter $K$ of MBA. The architecture is the one used in model $\#6$ from Table \ref{tab:ablation}.
		\texttt{(conv*)} means that we set a particular value of $K$ in that convolution stack only.}\label{tab:change_k}
	\smallskip
	\footnotesize{
		\begin{tabular}{ll}
			\hline
			\multicolumn{1}{l|}{Method}               & \multicolumn{1}{l}{Val. Error}  \\ \hline
			\multicolumn{1}{l|}{Baseline  }           & \multicolumn{1}{l}{9.4\%}     \\ \hline\hline
			\multicolumn{1}{l|}{$K=2$ }               & \multicolumn{1}{l}{8.9\%}    \\ \hline
			\multicolumn{1}{l|}{$K=4$ \texttt{(conv2)}, $K=2$ \texttt{(conv3)}}      & \multicolumn{1}{l}{8.1\%}    \\ 
			\multicolumn{1}{l|}{$K=4$}                  & \multicolumn{1}{l}{6.7\%}    \\ 
			\multicolumn{1}{l|}{$K=4$ \texttt{(fixed MBA)}}       & \multicolumn{1}{l}{10.8\%}    \\ \hline
			\multicolumn{1}{l|}{$K=8$}                  & \multicolumn{1}{l}{6.6\%}    \\ \hline
		\end{tabular}
	}
\end{table}

First, we explicitly explore the ability of the MBA module in CNN with different numbers of channels in each layer.
% and when MBA is applied to different number of layers. 
From Table \ref{tab:ablation} we conclude that
adding more MBA layers into the network generally reduces the classification error.
Also, the width of the network, \emph{i.e.}, the number of filters in each stack, plays an important role to reduce the classification error.
Considering models \texttt{\#4-\#6}, we can see that larger number of filters results in
more expressive power the network and thus smaller classification error.
%
%Model \#3 only uses MBA for the convolutional layers in stack 3 while model \#6 uses MBA for the convolutional layers in stacks 2 and 3. 
%
It can be observed that the use of more MBA layers in model \texttt{\#6} performs better than the use of fewer MBA layers in model \texttt{\#3}. 
However, the MBA module imposed on all stacks does not perform better than the one imposed on stack 2 and 3 only. 
This result shows that it is not necessary to use the MBA layer for lower convolutional layers that are close to raw data. 
Moreover, the improvement of our method does not come from introducing more parameters. 
For example, model \texttt{\#5} has much fewer parameters than the baseline 
%since the numbers of its filters have been reduced by half in the top two stacks. 
and it still outperforms the baseline.

Second, we investigate the effect of the hyperparameter $K$ in the MBA module (Table \ref{tab:change_k}), which is conducted on the CIFAR-10 validation set.
In the MBA layer, a channel is decoupled to $K$ channels.
We can observe that the inclusion of MBA layer on the network reduces the classification error when $K=2, 4$ and $8$.
To further validate the necessity of the learnable biases, we fix the bias parameters after initialization. 
In this case, the validation error increases from $6.7\%$ for learned bias to $10.8\%$ for fixed bias.
Moreover, we find that setting a large $K=8$ does not reduce the classification error further compared with $K=4$ because 
it is not necessary to decouple a single channel into too many channels.

\subsection{Comparison to State-of-the-Arts}
\label{Comparison to State-of-the-Arts}

\renewcommand{\arraystretch}{1.2}
\begin{table}[]
	\centering
	\caption{Classification test errors on CIFAR dataset with and without data augmentation. The best results are in bold.}
	\smallskip
	\label{my-label}
	\footnotesize{
		\begin{tabular}{lll}
			\hline
			\multicolumn{1}{l|}{Method}      & \multicolumn{1}{l|}{CIFAR-10} & CIFAR-100 \\ \hline
			\multicolumn{3}{c}{ \textit{Without Data Augmentation}   }    \\ \hline \hline
			\multicolumn{1}{l|}{ReLU \cite{dropout} }        & \multicolumn{1}{l|}{12.61\%}  & 37.20\%   \\ \hline
			\multicolumn{1}{l|}{Channel-out \cite{channel_out} } & \multicolumn{1}{l|}{13.20\%}  & 36.59\%   \\ \hline
			\multicolumn{1}{l|}{Maxout \cite{maxout}}       & \multicolumn{1}{l|}{11.68\%}  & 38.57\%   \\ \hline
			%\multicolumn{1}{l|}{Probout \cite{prob_out} }     & \multicolumn{1}{l|}{11.35\%}  & 38.14\%   \\ \hline
			\multicolumn{1}{l|}{NIN \cite{NIN}}         & \multicolumn{1}{l|}{10.41\%}  & 35.68\%   \\ \hline
			\multicolumn{1}{l|}{DSN \cite{DSN}}         & \multicolumn{1}{l|}{9.78\%}   & 34.57\%   \\ \hline
			\multicolumn{1}{l|}{APL \cite{piecewise_linear}} & \multicolumn{1}{l|}{9.59\%}   & 34.40\%   \\ \hline
			\multicolumn{1}{l|}{Ours}         & \multicolumn{1}{l|}{\textbf{6.73 \% }}   & \textbf{26.14\% }  \\ \hline	
			\multicolumn{3}{c}{
				%\smallskip 
				\textit{With Data Augmentation}} \\ \hline\hline
			\multicolumn{1}{l|}{Maxout \cite{maxout}}        & \multicolumn{1}{l|}{9.38\%}  & -   \\ \hline
			%\multicolumn{1}{l|}{Probout } & \multicolumn{1}{l|}{9.39\%}  & -   \\ \hline
			\multicolumn{1}{l|}{DropConnect \cite{drop_connect}}      & \multicolumn{1}{l|}{9.32\%}  & -   \\ \hline
			\multicolumn{1}{l|}{SelAtten \cite{sel_attention}}     & \multicolumn{1}{l|}{9.22\%}  & 33.78\%   \\ \hline
			\multicolumn{1}{l|}{NIN \cite{NIN}}         & \multicolumn{1}{l|}{8.81\%}  & -    \\ \hline
			\multicolumn{1}{l|}{DSN \cite{DSN}}         & \multicolumn{1}{l|}{8.22\%}   & -  \\ \hline
			\multicolumn{1}{l|}{APL \cite{piecewise_linear}} & \multicolumn{1}{l|}{7.51\%}   & 30.83\%   \\ \hline
			%\multicolumn{1}{l|}{AllCNN \cite{all_cnn}}         & \multicolumn{1}{l|}{7.25\%}   & 33.71\%   \\ \hline
			\multicolumn{1}{l|}{BayesNet \cite{scalable_bayes}}         & \multicolumn{1}{l|}{6.37\%}   & 27.40\%   \\ \hline
			\multicolumn{1}{l|}{ELU \cite{iclr16_elu}}         & \multicolumn{1}{l|}{6.55\%}   & 24.28\%   \\ \hline
			\multicolumn{1}{l|}{Ours}         & \multicolumn{1}{l|}{ \textbf{5.38\%} }   & \textbf{24.1\%} \\ \hline
		\end{tabular}
	}
\end{table}

We show the comparison results of the proposed MBA with other state-of-the-arts, including
ReLU \cite{dropout},
Channel-out \cite{channel_out},
Maxout \cite{maxout},
Network in Network (NIN) \cite{NIN},
Deep Supervision (DSN) \cite{DSN},
APL \cite{piecewise_linear},
DropConnect \cite{drop_connect},
Selective Attention Model \cite{sel_attention},
Scalable Bayes Network \cite{scalable_bayes} and
Exponential Linear Unit \cite{iclr16_elu}
on the CIFAR and SVHN datasets.
We will use the network architecture of candidate model \texttt{\#6} as the final MBA model thereafter.

\textbf{CIFAR.} 
Without data augmentation, 
Table \ref{my-label} indicates that we achieve a relative $29.8\%$ and $24\%$ gain over previous state-of-the-arts on CIFAR-10 and CIFAR-100, respectively.
As for the data augmentation version of our model, 
during trainin we first resize each image to a random size sampled from $[32, 40]$ and then crop a $32\times32$ region randomly out of
the resized image. Horizontal flip is also adopted. 
For testing, we employ a multi-crop voting scheme, where crops from five corners (center, top right and left, bottom right nad left) are extracted and the final score is determined by their average. 
Note that we do not aggressively resort to trying all kinds of data augmentation techniques \cite{scalable_bayes}, 
such as color channel shift, scalings, etc.; or extend the network's depth to extremely deep, 
for example, ELU \cite{iclr16_elu} used a model of 18 convolutional layers.
Our algorithm performs better than previous ones 
by an absolute reduction of $1.17\%$ and $0.18\%$ with data augmentation on these two datasets.

\renewcommand{\arraystretch}{1.2}
\begin{table}[]
	\centering
	\caption{Classification test errors on SVHN dataset without data augmentation. The best results are in bold.}
	\smallskip
	\footnotesize{
	\begin{tabular}{l | l}
		\hline
		Method  & Test Error  \\ \hline
	      StoPool \cite{stochastic_pool} & 2.80\% \\ \hline
	     ReLU \cite{dropout} & 2.55\% \\ \hline
	     Maxout \cite{maxout}  & 2.47\%   \\ \hline
	     NIN \cite{NIN} &  2.35\% \\ \hline
	     DropConnect \cite{drop_connect} & 1.94\% \\ \hline
	     DSN \cite{DSN}  & 1.92\%  \\ \hline
	     GenPool \cite{gen_pool} & \textbf{1.69\%} \\ \hline
	     Ours    &  1.80\%   \\ \hline	     
	    \end{tabular}\label{svhn}
	    \vspace{-.5cm}
}
\end{table}

\textbf{SVHN.} We further conduct the comparison experiment on the house number dataset and we achieve
a test error rate of 1.80\% without data augmentation (Table \ref{svhn}).

\section{Conclusion and Discussion}\label{conclusion}

In this work, we propose a multi-bias non-linearity activation (MBA) module in deep neural networks.
A key observation is that magnitudes of the responses from convolutional kernels have a wide diversity of pattern representations in the network, 
and it is not proper to discard weaker signals with single thresholding.
The MBA unit placed after the feature maps helps to decouple response magnitudes to multiple maps and
 generates more patterns in the feature space at a low computational cost.
We demonstrate that our algorithm is effective by conducting various independent 
component analysis as well as comparing
the MBA method with other state-of-the-art network designs.
Experiments show that such a design has superior performance than previous state-of-the-arts.

While the MBA layer could enrich the expressive power of the network, we believe more exploration of
the discriminative features  can be investigated to leverage the information hidden in the magnitude of response.
Such an intuition is triggered by the fact that the non-linearity actually preserves or maintains the depth property of a network.
One simple way is to divide the input feature maps, feed them into multiple non-linearities, and gather together again as the input of the subsequent 
convolutional layer.
%We will leave this point as our future work.

\bibliographystyle{icml2016}
\bibliography{deep_learning}

\end{document}